%% file: example_paper.tex
\documentclass{article}

\usepackage{microtype}
\usepackage{graphicx}
\usepackage{subfigure}
\usepackage{booktabs} 

\usepackage{hyperref}

\usepackage[accepted]{icml2024}

\usepackage{amsmath}
\usepackage{amssymb}
\usepackage{mathtools}
\usepackage{amsthm}

\usepackage[capitalize,noabbrev]{cleveref}

\theoremstyle{plain}

\theoremstyle{definition}

\theoremstyle{remark}

\usepackage{pifont}
\usepackage{tabularray}
\usepackage{amssymb}
\usepackage{multirow}
\usepackage{cleveref}

\definecolor{forestgreen}{rgb}{0.13, 0.55, 0.13}
\usepackage{color, colortbl}
\definecolor{azure}{rgb}{0.0, 0.5, 1.0}
\definecolor{neutral_gray}{HTML}{D0CDCD}
\definecolor{challenge_yellow}{HTML}{FFD451}
\definecolor{darkgreen}{HTML}{448E64}
\definecolor{neg_red}{HTML}{FFB0BA}
\definecolor{pos_green}{HTML}{CAECD0}
\definecolor{applegreen}{rgb}{0.55, 0.71, 0.0}
\definecolor{success}{HTML}{5CB85C}
\definecolor{tableblue}{HTML}{D0EFFF}
\definecolor{tablepink}{HTML}{FBE5D6}
\usepackage{amssymb}
\usepackage{pifont}
\newcommand{\xmark}{\ding{55}}%
\usepackage{color}

\DeclareRobustCommand{\hlc}[1]{{\sethlcolor{challenge_yellow}\hl{#1}}}
\DeclareRobustCommand{\hlp}[1]{{\sethlcolor{pos_green}\hl{#1}}}
\DeclareRobustCommand{\hln}[1]{{\sethlcolor{neg_red}\hl{#1}}}
\usepackage{soul}

\usepackage[textsize=tiny]{todonotes}

\icmltitlerunning{Reliable, Adaptable, and Attributable Language Models with Retrieval}

\begin{document}

\twocolumn[
\icmltitle{
Reliable, Adaptable, and Attributable Language Models with Retrieval}



\icmlsetsymbol{equal}{*}

\begin{icmlauthorlist}
\icmlauthor{Akari Asai}{uw,meta}
\icmlauthor{Zexuan Zhong}{pri}
\icmlauthor{Danqi Chen}{pri}
\icmlauthor{Pang Wei Koh}{uw}\\
\icmlauthor{Luke Zettlemoyer}{uw,meta}
\icmlauthor{Hannaneh Hajishirzi}{uw,ai2}
\icmlauthor{Wen-tau Yih}{meta}
\end{icmlauthorlist}

\icmlaffiliation{uw}{University of Washington}
\icmlaffiliation{pri}{Princeton University}
\icmlaffiliation{meta}{Meta AI}
\icmlaffiliation{ai2}{Allen Institute for AI}

\icmlcorrespondingauthor{Akari Asai}{akari@cs.washington.edu}

\icmlkeywords{Machine Learning, ICML}

\vskip 0.3in
]



\printAffiliationsAndNotice{} 

\begin{abstract}
Parametric language models (LMs), which are trained on vast amounts of web data, exhibit remarkable flexibility and capability. 
However, they still face practical challenges such as hallucinations, difficulty in adapting to new data distributions, and a lack of verifiability. 
In this position paper, we advocate for retrieval-augmented LMs to replace parametric LMs as the next generation of LMs. By incorporating large-scale datastores during inference, retrieval-augmented LMs can be more reliable, adaptable, and attributable. 
Despite their potential, retrieval-augmented LMs have yet to be widely adopted due to several obstacles:  
specifically, current retrieval-augmented LMs 
struggle to leverage helpful text beyond knowledge-intensive tasks such as question answering, 
have limited interaction between retrieval and LM components, and lack the infrastructure for scaling. 
{To address these, we propose a roadmap for developing general-purpose retrieval-augmented LMs. 
This involves a reconsideration of datastores and retrievers, the exploration of pipelines with improved retriever-LM interaction, and significant investment in infrastructure for efficient training and inference. 
}

\end{abstract}

\section{Introduction}
\label{sec:intro}
\input{sections/sec1_intro}

\section{How Far Can We Go with Parametric LMs?}
\label{sec:parametric}
\input{sections/sec2_parametic}

\section{How Can Retrieval-Augmented LMs Address These Issues?}
\label{sec:ret_lm}
\input{sections/sec3_semi_parametric}

\section{How Can We  Further Advance Retrieval-Augmented LMs? }
\label{sec:challenges}
\input{sections/sec4_challenges}

\section{Conclusion}
\label{sec:conclusion}
\input{sections/sec5_conclusion}

\section*{Acknowledgements}
We express our gratitude to Jacqueline He for her meticulous proofreading and for providing many writing suggestions for the drafts. 
We thank Rulin Shao, Weijia Shi, Dan Friedman, Tanya Goyal, Howard Yen, Dhruba Ghosh, Jiacheng Liu, and Niklas Muennighoff for their insightful feedback on our draft, and Sewon Min for fruitful discussions in the early stages. 
This work was funded in part by NSF IIS-2044660 and IIS-2239290 and gifts from AI2. 

\bibliography{example_paper}
\bibliographystyle{icml2024}

\newpage
\appendix
\onecolumn
\input{sections/appendix}

\end{document}

%% file: sections/sec1_intro.tex
\begin{figure}[t]
    \centering
    \includegraphics[width=0.95\linewidth]{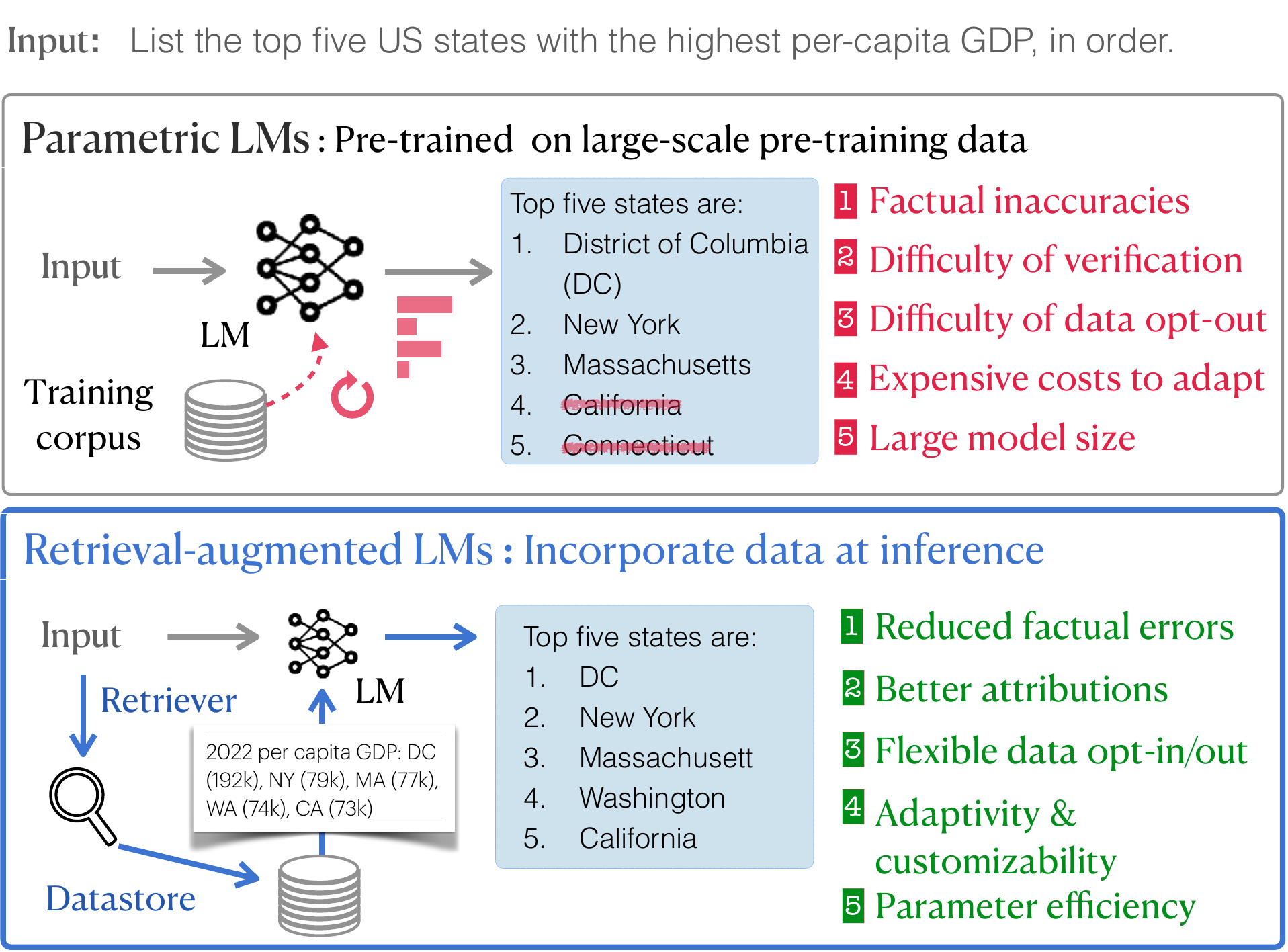}
    \caption{
{Parametric LMs (top) internalize large-scale text data in their parameters via massive pre-training, while retrieval-augmented LMs (bottom) incorporate text retrieved from a massive {\it datastore} at test time.}
    }
    \label{fig:fig1}
\end{figure}

Large language models (LMs) such as GPT-4~\cite{black2022gpt} have shown impressive abilities in a range of natural language processing (NLP) tasks.
Such {\bf parametric LMs}
encapsulate rich natural language understanding abilities and a wealth of world knowledge in their parameters, acquired via massive pre-training on large-scale web corpora (Figure~\ref{fig:fig1}, top). 
However, they still suffer from several fundamental weaknesses including \hln{\bf W1}: the prevalence of factual errors~\cite{min2023factscore,mishra2024finegrained}, \hln{\bf W2}: the difficulty of verification~\cite{bohnet2022attributed},  \hln{\bf W3}: {difficulty of opting out certain sequences with concerns}~\cite{henderson2023foundation}, \hln{\bf W4}: computationally expensive costs for adaptations~\cite{longpre2023pretrainer}, and \hln{\bf W5}: prohibitively large model size~\cite{kandpal2022large}. 
Moreover, merely scaling up the model has been insufficient to overcome such limitations~\cite{mallen2022not} or even exacerbates the challenges~\cite{carlini2021extracting}.

This position paper advocates for retrieval-augmented LMs to supersede parametric LMs as the next generation of LMs (Figure~\ref{fig:fig1}, bottom), addressing many of the aforementioned weaknesses. 
Unlike parametric LMs---which use large-scale text data only during training---retrieval-augmented LMs leverage an external large-scale collection of documents ({\it datastore}) at inference by selecting relevant documents from the datastore~\cite{asai-etal-2023-retrieval}. 
Retrieval-augmented LMs can \hlp{\bf W1}: largely reduce factual errors~\cite{mallen2022not}, \hlp{\bf W2}: provide better attributions \cite{gao2023enabling}, {\hlp{\bf W3}: {enabling flexible opt-in and out of sequences}~\cite{min2023silo}}. 
By adding or removing data from their datastores, retrieval-augmented LMs can \hlp{\bf W4}: easily adapt to new  distributions~\cite{khandelwal2019generalization}. 
Lifting the burden of memorizing everything in parameters makes them \hlp{\bf W5}: more parameter-efficient~\cite{izacard2022few}. 

Despite their considerable potential to significantly improve reliability, adaptability, and attributability, their broader adoption beyond specific knowledge-intensive tasks (e.g., question answering or QA; \citealt{chen-etal-2017-reading}) is currently limited. 
We argue that through fundamental advancements in architecture, training methodologies, and infrastructure for retrieval-augmented LMs, they can demonstrate substantial efficacy across diverse domains. 
We urge the research community to intensify efforts aimed at overcoming those inherent limitations for their widespread adoption. 
To facilitate future research, we identify several significant challenges.
First, existing approaches primarily leverage context with high semantic or lexical similarity to the input (\hlc{\bf C1}), struggling when helpful text is absent in common datastores or does not align with conventional relevance definitions~\cite{behnamghader-etal-2023-retriever, asai-etal-2023-task}. 
Second, prepending the retrieved text to the input of off-the-shelf LMs, which has been widely used recently, leads to shallow interactions between the retrieval and LM components (\hlc{\bf C2}). This often results in unsupported generations~\cite{gao2023enabling}, susceptibility to irrelevant text~\cite{Yoran2023MakingRL}, and challenges in handling information from multiple pieces of  text~\cite{borgeaud2021improving}. 
{Furthermore, unlike rapid progress for efficient training and inference of parametric LMs~\cite{10.14778/3611540.3611569,dao2022flashattention}, there are limited studies and open-sourced efforts to enhance the training and inference efficiency of retrieval-augmented LMs at scale (\hlc{\bf C3}).

{We conclude this paper with a roadmap to advance retrieval-augmented LMs to foster wider adoption.} 
First, addressing the challenge of finding helpful text for diverse tasks (\hlc{\bf C1}), it is important to reconsider the notion of relevance and advance our understanding of what constitutes an effective datastore---specifically, exploring the types of information that should be retrieved from various datastores to enhance the performance in broader tasks.
Then, we suggest approaches to ensure deeper interactions between the two components, including architecture, pre-training, and post-training adaptations (\hlc{\bf C2}), rather than focusing on supplementary enhancement of existing parametric LMs. 
For challenges of scaling (\hlc{\bf C3}), we call for more open-sourced and interdisciplinary efforts across hardware, systems, and algorithms to develop infrastructures for training and inference (e.g., scaling datastore to trillion tokens).
By pursuing these avenues, we anticipate unlocking the full capabilities of retrieval-augmented LMs and expanding their applications across a spectrum of tasks and domains.

%% file: sections/sec2_parametic.tex
We first assess the limitations of parametric LMs. 
Despite rapid progress in this area, we argue that parametric LMs have many practical limitations, which in turn pose significant challenges to building reliable intelligent systems. 

\paragraph{Definition.}
~A parametric LM (Figure~\ref {fig:fig1}, top) consists of a set of parameters $\theta$. 
Given input sequences from a large-scale text dataset $\mathcal{D}_{\rm train}$, learnable parameters $\theta$ are trained to predict the probabilities of future or masked tokens.  
During test time, for an input sequence $x$, the trained $\theta$ predicts the outputs: $ y =f_\theta(x)$, without accessing any external data beyond that of the task at hand. 

\subsection{Weaknesses of Parametric LMs}
\label{sec:limitations_paarmetric}
Mounting evidence highlights significant limitations in parametric LMs. Many such challenges arise from the strategy of attempting to store all knowledge within the parameters, which scaling alone may not adequately address. 

\vspace{0.1cm}
\noindent{\bf \hln{W1: Factual inaccuracies}. }
~Attempting to memorize all the learned knowledge within the parameters can lead to factual inaccuracies, which are often called hallucinations. 
Several recent papers report that even state-of-the-art LMs such as ChatGPT exhibit hallucinations in the majority of their outputs~\cite{min2023factscore,mishra2024finegrained}. 
\citet{mallen2022not,kandpal2022large} show that they particularly struggle with long-tail knowledge---factual knowledge that is less represented during pre-training---and that scaling only yields minor improvements.  
\citet{gudibande2023false} find that increasing synthetic labeled data during instruction tuning may not improve the factuality of model outputs. 

\paragraph{\hln{W2: Difficulty of verifications}. }
Not only have LMs shown a propensity for hallucinations in their generations, but it is also difficult for practitioners to fact-check their outputs due to a lack of clear attributions or provenance. 
The outputs of powerful LMs are often lengthy, assertive, and plausible~\cite{min2023factscore}, which makes post-hoc attributions or factual verification to be challenging and largely unsolved tasks~\cite{mishra2024finegrained,yue2023automatic}.   

\paragraph{\hln{W3: Difficulty of opting out certain sequences from the datasets}.}
{
Managing the vast volume of pre-training data poses a considerable challenge in identifying and filtering out training instances with potential privacy~\cite{brown2022does} or copyright-protected data~\cite{lee2023talkin}. 
Recent work studies intensive red teaming and safety tuning efforts~\cite{touvron2023llama2,perez2022red}, unlearning~\cite{jang-etal-2023-knowledge} or iterative pre-training of models on corpora after removing certain data~\cite{kandpal2022deduplicating}. 
Yet, the absence of proper attributions further complicates these endeavors, as tracing back to and eliminating specific training instances becomes non-trivial~\cite{grosse2023studying}. 
}

\paragraph{\hln{W4: Computationally expensive costs to adapt}.} 
Adapting parametric LMs trained on static unlabeled text (i.e., text collected at a certain timestamp from the web) requires continuous training or computationally expensive post-adaptation to new data distributions. 
For instance, their parametric knowledge can quickly become obsolete~\cite{longpre2023pretrainer}. 
While several approaches propose to locate and edit certain outdated knowledge~\cite{de-cao-etal-2021-editing} or conduct efficient continued training~\cite{jin-etal-2022-lifelong-pretraining} to keep up with the world, these approaches require additional computationally expensive learning processes. 
LMs trained on widely adopted pre-training corpora often perform well on general-purpose domains such as news articles~\cite{dodge-etal-2021-documenting}, but struggle on expert domains~\cite{taylor2022galactica}. 
Prior work demonstrates the effectiveness of continued pre-training~\cite{azerbayev2023llemma,chen2023meditron} or instruction tuning~\cite{singhal2023large}, albeit at a considerable computational cost {and possibilities of catastrophic forgetting~\cite{li-etal-2022-overcoming}}.

\vspace{-0.2cm}
\paragraph{\hln{W5: Prohibitively large model size}. }
Numerous studies showcase the positive impact of model scaling on task performance~\cite{chowdhery2022palm,wei2022emergent}, and the ability to recall factual knowledge memorized from the training data~\cite{Carlini2022QuantifyingMA,mallen2022not,kandpal2022large}. 
This trend has prompted the community to focus on boosting the model size in pursuit of better performance, at the cost of significant computational challenges and environmental concerns~\cite{strubell-etal-2019-energy,weidinger2022taxonomy}. 
Despite efforts to enhance efficiency, hosting these massive models, which often exceed a hundred billion parameters, remains impractical for many industry or academic groups~\cite{schwartz2019green}.

%% file: sections/sec3_semi_parametric.tex
In this section, we discuss how retrieval-augmented LMs can alleviate the aforementioned issues in parametric LMs. 

\begin{figure*}[t!]
    \centering
    \includegraphics[width=0.99\linewidth]{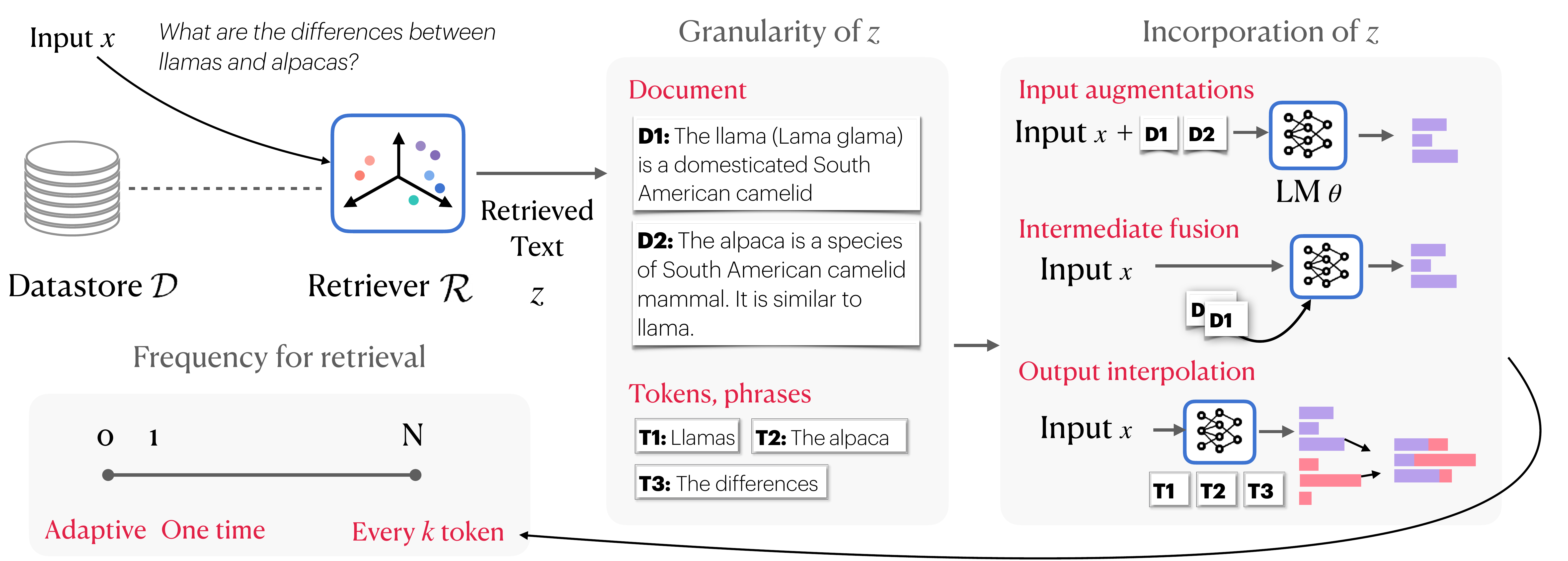}
    \caption{Taxonomy of architectures of retrieval-augmented LMs. 
    }
    \label{fig:fig2}
\end{figure*}

\vspace{0.1cm}
\noindent{\bf Definition. }
~A retrieval-augmented LM (Figure~\ref {fig:fig1}, bottom; detailed in Figure~\ref{fig:fig2}) typically consists of two key components: a retriever $\mathcal{R}$ and a parametric LM $\theta$. 
The retriever builds a search index $\mathcal{I}$\footnote{In term-based retrieval systems such as BM25~\cite{Robertson2009ThePR} that count the occurrences of words in documents in the datastore, the index $\mathcal{I}$ is a weighted bag-of-words vector, while in more recent trainable neural retrieval systems such as DPR~\cite{karpukhin2020dense}, the index is a collection of float embeddings encoded by an encoder LM. }   based on documents in the datastore $\mathcal{D}$.
During inference time, given an input sequence $x$, the retriever finds relevant text $z$\footnote{There are different granularities for relevant text $z$ (e.g., text chunks, tokens, phrases). See Section~\ref{sec:survey_arc} for more details.} from the inference datastore, leveraging an index $\mathcal{I}$: $z = f_{\mathcal{R},\mathcal{I}}(x)$. 
Subsequently, the LM $\theta$ uses both the original prompt and the retrieved text to predict the output $y$: $y = f_{\theta}(x, z)$.  

\vspace{0.1cm}
\noindent{\bf Origins, progress, and recent shift. }
~The concept of retrieval augmentation has been extensively explored across various machine learning domains~\cite{tian2018learning}. 
In NLP, earlier efforts have been applied to specific tasks such as QA and machine translation. 
\citet{chen-etal-2017-reading} introduce DrQA, which combines a term-based information retrieval (IR) system with a neural QA model to answer knowledge-intensive questions. 
While IR and such task LMs were initially studied separately, several work explores more organic combinations of retrieval and LM by pre-training the two components jointly or sequentially, including REALM~\cite{guu2020retrieval}, RAG~\cite{lewis2019bart}, RETRO~\cite{borgeaud2021improving}, {\it etc}.

Such earlier work designed special architectures and training objectives for the retrieval-augmented LM. 
Most recently, there has been a shift of view of retrieval-augmented LMs---instead of training retrieval-augmented LMs from scratch, some work supplementary integrate retrieval on top of existing powerful parametric LMs (e.g., GPT-3; \citealt{black2022gpt}) without any additional training. 
Such methods---often referred to simply as Retrieval-Augmented Generation or RAG---concatenate the original input sequence $x$ with retrieved text $z$ when prompting, yielding significant improvements over the base parametric LMs on certain knowledge-intensive tasks~\cite{ram2023context,shi2023replug}. 
Many recent studies explore advanced prompting methods with retrieval components~\cite{yao2023react,press-etal-2023-measuring} or develop pipelines for further improvements~\cite{gao2023retrieval}. 
RAG has been integrated into real-world applications such as LLM search systems.\footnote{\url{https://bard.google.com/chat}}

\subsection{Effectiveness of Retrieval-Augmented LMs}
\label{sec:effectiveness}

We now review some empirical findings from prior studies suggesting their effectiveness in addressing the weaknesses of parametric LMs discussed in Section~\ref{sec:limitations_paarmetric}. 

\paragraph{\hlp{W1: Reduced factual errors in long-tail knowledge}.}
Recent studies show that retrieval-augmented LMs can alleviate the shortcomings of parametric memorization by explicitly capturing long-tail knowledge~\cite{mallen2022not}. As a result, retrieval-augmented LMs can minimize hallucinations and improve the factuality of generated outputs~\cite{lewis2020retrieval,izacard2022few,ram2023context,shi2023replug,asai2023self,min2022nonparametric}.
 
\vspace{-0.3cm}
\paragraph{\hlp{W2: Better attributions}.}
Retrieval-augmented LMs provide retrieved results $z$ used during inference, which can help practitioners inspect the correctness of model outputs manually~\cite{liu-etal:2023:tacl} or automatically~\cite{mishra2024finegrained}.
{Another way for verification is post-hoc attribution---given the model output $y$, retrieving documents that support $y$.  
Yet, prior work finds that retrieval-augmented LMs using evidence during inference provide more accurate attributions than such post-hoc attributions~\cite{gao2023enabling,Malaviya2023ExpertQAEQ}
}

\paragraph{\hlp{W3: Enabling flexible opt-in of sequences}.}
{
Retrieval-augmented LMs offer some effective solutions to concerns related to massive training data through improved attributions and adaptable datastore updates.
Enhanced attributions enable practitioners to exclude specific sequences from the datastore, mitigating the risk of generating them verbatim~\cite{carlini2021extracting}. 
Additionally, integrating datastores during inference only still allows retrieval-augmented LMs to maintain performance across domains not included in their training data~\cite{min2023silo}. 
}

\paragraph{\hlp{W4: Adaptability and customizability}.}
The separation and the interchangeability of knowledge sources for the datastore enables better customization to specific domains, applications, and time stamps, without the need for additional training~\cite{khandelwal2019generalization,min2023silo}. 
Recent work has shown that retrieval augmentation can even outperform LMs fine-tuned on the downstream domain data on QA~\cite{ovadia2023fine,gupta2024rag}.
Such effectiveness for domain adaptation has also been reported in non-knowledge-intensive tasks, including machine translation~\cite{shi2022nearest,min2023silo,khandelwal2021nearest,zhong2022training}.
Updating the datastore with up-to-date knowledge also bypasses the issue of data obsoleteness of parametric LMs~\cite{izacard2022few,zhong-etal-2023-mquake,pmlr-v162-mitchell22a,kasai2022realtime}.

\paragraph{\hlp{W5: Parameter efficiency}.}
By lifting the burden of memorizing all knowledge in the model parameters, retrieval-augmented LMs often show strong parameter efficiency---retrieval-augmented LMs with much fewer LM parameters can outperform larger, more powerful parametric LMs. 
For example, on knowledge-intensive tasks such as QA, retrieval-augmented LMs surpass parametric LMs with orders of magnitude more parameters by a large margin~\cite{izacard2022few,min2022nonparametric,mallen2022not}.

\begin{table*}[]
    \centering
    \footnotesize
\caption{Diverse retrieval-augmented LMs based on our architecture and training taxonomies. Full references of the papers are as follows: DrQA~\cite{chen-etal-2017-reading}, REALM~\cite{guu2020retrieval}, RAG~\cite{lewis2020retrieval}, ATLAS~\cite{izacard2022few}, RALM~\cite{ram2023context}, REPLUG~\cite{shi2023replug}, Active Retriever~\cite{jiang2023active}, Self-RAG~\cite{asai2023self}, RETRO~\cite{borgeaud2021improving}, InstructRetro~\cite{anonymous2023instructretro}, kNN LM~\cite{khandelwal2019generalization}, TRIME~\cite{zhong2022training}, NPM~\cite{min2022nonparametric}, CopyGenerator~\cite{Lan2023CopyIA}, SPALM~\cite{10.1162/tacl_a_00371}, Adaptive kNN~\cite{drozdov-etal-2022-cant}. {$^*$ indicates that approaches combining off-the-shelf models without any task-specific training.} 
}
    \label{tab:taxonomy}
    \begin{tabular}{l|lll|l|l}
    \toprule
    &   Granularity & Incorporation & Frequency &  Training & Data order \\\midrule
    DrQA &  Chunks & Input & One-time & Independent & $O(10^9)$ \\
    REALM, RAG, ATLAS &  Chunks & Input & One-time & Joint & $O(10^9)$  \\
    RALM, REPLUG &  Chunks & Input & Every $k$ tokens, One-time & Independent$^*$ & $O(10^9)$  \\
    Active-Retriever, Self-RAG &  Chunks & Input &  Adaptive & Independent$^*$, Sequential & $O(10^9)$  \\
    RETRO, InstructRetro &  Chunks & Intermediate &  Every $k$ tokens & Sequential & $O(10^{12})$  \\
    kNN LM, TRIME &  Tokens & Output &  Every token & Independent$^*$, Joint & $O(10^{9})$  \\
    NPM, Copy Generator &  Phrases & Output &  Every phrase & Joint & $O(10^{9})$  \\
    SPALM, Adaptive kNN &  Tokens & Output &  Adaptive & Independent$^*$, Joint & $O(10^{9})$  \\
\bottomrule
    \end{tabular}
\end{table*}

\section{Why Haven't Retrieval-Augmented LMs Been Widely Adopted?}
\label{sec:current_state}
Despite showing some empirical promise, the adoption of retrieval-augmented LMs remains limited compared to parametric LMs. 
To understand the obstacles hindering the widespread adoption, we provide a brief review of existing retrieval-augmented LMs under our unified taxonomy for architectures (Figure~\ref{fig:fig2}), training, and datastores, as summarized in Table~\ref{tab:taxonomy}.

\subsection{Current State of Retrieval-Augmented LMs}
\label{sec:survey} 

\subsubsection{Architecture}
\label{sec:survey_arc}
{Retrieval-augmented LMs have diverse architectures. Our taxonomy defines architecture based on three axes (Table~\ref{tab:taxonomy} left): what the unit of retrieved text $z$ is (granularity of $z$), how $z$ is incorporated (incorporation of $z$), and how often $z$ is retrieved (frequency of retrieval). }

Here, we classify approaches based on how they incorporate the retrieved text $z$ {(the {Incorporation} column in Table~\ref{tab:taxonomy})}, 
Essentially, retrieval-augmented LMs' architectures can be classified into the following three groups: 1)  {\bf input augmentation}, 2) {\bf intermediate fusion}, and 3) {\bf output interpolation}. Refer to Figure~\ref{fig:fig2} for a taxonomy of these architectures. 
For a more comprehensive review of the architecture, including aspects such as the granularity of retrieval and retrieval frequency, refer to Appendix~\ref{app_sec_survey_arc}. 
In essence, input augmentation and intermediate fusion typically involve retrieving text chunks and processing them with parametric LMs. 
On the other hand, output interpolation directly retrieves successive tokens or phrases, resulting in a much larger index. 
Unlike traditional approaches, which involve retrieving only once {(One-time)} such as DRQA, recent studies have highlighted the effectiveness of retrieval over specific token intervals {(Every $k$ tokens; ~\citealt{ram2023context})} or adaptively~\cite{asai2023self,jiang2023active,drozdov-etal-2022-cant}. 

\vspace{0.1cm}
{\bf Input augmentation.}
Input augmentation augments the original input $x$ with retrieved results $z$ in the input space of the LM $\theta$ and runs a standard LM inference. 
As in the pioneering work from \citet{chen-etal-2017-reading}, input augmentation enables flexible plug-ins of different models for retrieval and LM components. 
Many widely adopted models, including those that augment powerful LMs with off-the-shelf retrievers, mostly belong in this category~\cite{yao2023react,shi2023context}.   
One notable bottleneck to this approach is redundancy and inefficiency; encoding many documents together in the input space leads to context length window limitations and increases inference costs exponentially~\cite{xu2023recomp}. 
While some work such as FiD~\cite{izacard2022few} 
explores parallel encoding to overcome such inefficiencies, it still encodes repeatedly for each input $x$. 

\vspace{0.1cm}
{\bf  Intermediate fusion.} 
To integrate retrieved results in a more scalable manner, RETRO~\cite{borgeaud2021improving} introduces a new attention mechanism, which takes many pre-encoded text chunks independent of query $x$ and simultaneously incorporates them in intermediate spaces. 
RETRO++~\cite{wang-etal-2023-shall} and InstructRetro~\cite{anonymous2023instructretro} demonstrate the effectiveness of this method on top of larger, decoder-only LMs.  
However, a drawback of intermediate fusion is the need for extensive architecture modification and pre-training of LMs for the new encoding blocks, potentially limiting widespread adoption. 

\vspace{-0.1cm}
\paragraph{ Output interpolation.}
Both input augmentation and intermediate fusion require the LM to generate continuations from their vocabularies.
In contrast, kNN LM~\cite{khandelwal2019generalization} interpolates a parametric LM token distribution with a retrieved token distribution, without the need for additional training. 
Some work extends this direction by designing new training objectives~\cite{zhong2022training} or completely replacing parametric distributions with a non-parametric distribution over each phrase in the datastore~\cite{min2022nonparametric,Lan2023CopyIA}.b
While these approaches frequently demonstrate their efficacy compared to input augmentation in language modeling ~\cite{min2023silo}, they require a considerably larger index than the other two architectures. This is due to the necessity of generating embeddings for all tokens in the datastore, presenting scalability challenges.

\subsubsection{Training}
\label{sec:survey_training}
Retrieval-augmented LMs consist of three main components: the index $\mathcal{I}$, the retriever $\mathcal{R}$ {(i.e., a model that generates encoding of input and documents)}, and the LM $\theta$. 
How to efficiently and simultaneously update them to optimize the whole pipeline remains a challenging question. 
Currently, there are two paradigms: {\bf independent or sequential training} and {\bf joint training} (Table~\ref{tab:taxonomy} Training). 

\vspace{-0.2cm}
\paragraph{Independent or sequential training.}
Independent training involves the separate development of a retriever and LM with no direct interactions during training. 
This includes methods such as kNN LM, or recently, RAG applied to off-the-shelf LMs and retrieval systems.  
This allows practitioners to leverage existing training pipelines and objectives to enhance the individual components. 
There has been rich literature in the area of IR on how to build reliable and efficient IR systems. 
Classical term-based retrieval systems, such as TF-IDF or BM25~\cite{Robertson2009ThePR}, have been widely used. 
More recently, neural retrieval systems, such as DPR~\cite{karpukhin2020dense} or ColBERT~\cite{khattab2020colbert}, 
have shown superior performance. 
Extensive pre-training of retrieval models further improved such models~\cite{izacard2021towards,ni-etal-2022-large,lin-etal-2023-train}.  
For a comprehensive review of retrieval systems, we direct readers to prior surveys~\cite{10.1145/3637870}.

Yet, independent training is often sub-optimal for the whole retrieval-augmented LM pipeline; for instance, LMs trained without retrieval could become easily distracted by irrelevant preceding context~\cite{shi2023large}. 
To alleviate this issue, sequential training trains either the retriever or LM first, and then trains the other subsequently using signals from the first trained component. 
Many studies train the LM component with a powerful pre-trained retriever e.g., DPR, search engines, or frozen pre-trained encoders~\cite{izacard-grave-2021-leveraging,nakano2021webgpt,borgeaud2021improving}, or conversely, train the retriever with signals from the LM~\cite{shi2023replug,izacard2021distilling}. 

\vspace{0.2cm}
\noindent{\bf Joint training. }
Joint training simultaneously trains the LM and retrieval components to further optimize their interactions and the end-to-end retrieval-augmented LM pipeline. 
{A notable challenge in joint training is the substantial computational overhead incurred by updating both the retriever model and the resulting index during training. It is impractical to repeatedly generate embeddings for millions or billions of documents in the datastore at each time step. }
There are two approaches to achieve this under reasonable resource requirements: updating the datastore with updated parameters asynchronously or using an in-batch approximation to a full datastore.  
{\bf Asynchronous updating} is a technique that allows the index to grow stale over a fixed number of training steps before the update, aiming to use the full corpus during training~\cite{izacard2022few}{, as in inference time. }
There is a tradeoff between the update frequency and computational overhead~\cite{guu2020retrieval}: to obtain better performance, the index should be updated more frequently. 
{\bf In-batch approximation} builds a temporary index on the fly using training samples from the same mini-batch, which serves as an approximation to the full index during training~\cite{zhong2022training,de2021mention, min2022nonparametric,Lan2023CopyIA}.
Designing training batches that can provide strong training signals requires careful consideration. 

\begin{table*}[]
    \centering
    \small
    \renewcommand{\arraystretch}{1.1}
\caption{Current status of retrieval-augmented LMs and future directions.  
}
    \label{tab:current_status}
    \begin{tabular}{p{2.2cm} |p{6.5cm}|p{7cm}}
    \toprule
    &  Current State of Retrieval-Augmented LMs (\S\ref{sec:current_state}) & Roadmap to Advance Retrieval-Augmented LMs (\S\ref{sec:challenges})  \\ \midrule
    \hlc{C1}: Usage of $\mathcal{R}$ & \textcolor{red}{\bf \xmark} Semantic and lexical similarity only & \textcolor{darkgreen}{\bf \checkmark} Beyond semantic and lexical similarity \\
    and $\mathcal{D}$ & \textcolor{red}{\bf \xmark} Single and general-domain corpora & \textcolor{darkgreen}{\bf \checkmark} Datastores for wider applications \\\hline
    \hlc{C2}: Interaction  & \textcolor{red}{\bf \xmark} Limited interactions beyond input augmentations  & \textcolor{darkgreen}{\bf \checkmark} Architectures with deep  LM-retriever interactions \\
    of $\mathcal{R}$  and $\mathcal{\theta}$ & \textcolor{red}{\bf \xmark} Lack of joint optimization from the end use & \textcolor{darkgreen}{\bf \checkmark} Large-scale joint training techniques\\\hline
    \hlc{C3}: Infrastructures for scaling &  \textcolor{red}{\bf \xmark} Lack of standardized libraries beyond RAG &{\textcolor{darkgreen}{\bf \checkmark} Standardized and open-sourced library for retrieval-based LMs} \\
      \& adoptions  & \textcolor{red}{\bf \xmark} Difficulty in large-scale training and inference & \textcolor{darkgreen}{\bf \checkmark} Infrastructure for large-scale training and inference \\
\bottomrule
    \end{tabular}
\end{table*}

\subsubsection{Applications and Datastores}
\label{sec:survey_applications}
\paragraph{Applications.}
Retrieval-augmented LMs have proven effective in various NLP tasks. 
Notably, their impact is more pronounced on knowledge-intensive tasks~\cite{guu2020retrieval,lewis2019bart,izacard2022few}. 
Several studies showcase their efficacy in machine translation~\cite{khandelwal2019generalization,Gu2017SearchEG} as well as broader language understanding tasks~\cite{min2022nonparametric,shi2022nearest}.
There are also decoding methods that leverage post-hoc retrieval augmentations to produce more efficient or factual generations~\cite{he2023rest,shi2023trusting}, or knowledge editing capabilities~\cite{zhong-etal-2023-mquake}. 
A further overview of applications with details of adaptation methodologies is in Appendix~\ref{app_sec:applications_datastore}.

\vspace{0.2cm}
{\bf Datastores.}~
Designing and building a reliable datastore is a key challenge of retrieval-augmented LMs. 
The inference datastore $\mathcal{D}$ may not be necessarily equivalent to the training datastore and is task-dependent. 
Some works, such as NPM~\cite{min2022nonparametric}, leverage the same corpus as the training data $\mathcal{D} = \mathcal{D}_{\rm train}$ on more general tasks, while for certain downstream tasks, a smaller and general-domain corpus is often used (e.g., Wikipedia).  
Conversely, curating high-quality, domain-focused corpora is important for some tasks, e.g., code generation~\cite{hayati-etal-2018-retrieval,zhou2023docprompting}. 
As Table~\ref{tab:taxonomy} shows, most prior work use a datastore that is on the order of $O(10^{9})$ tokens, with examples such as Wikipedia containing roughly a few billion tokens. 
Notably, \citet{anonymous2023instructretro,borgeaud2021improving} scale the datastore to over one trillion tokens, showcasing a large perplexity reduction.

\subsection{Limitations of Current Retrieval-Augmented LMs}
\label{sec:limitations}

We next identify several core challenges inherent to existing retrieval-augmented LMs, as summarized in Table~\ref{tab:current_status}. 

\paragraph{\hlc{C1: Limitations of retrievers and datastores}.}
Despite the success of retrieval-augmented LMs on knowledge-intensive tasks, their broader applications often result in restricted success. 
For example, retrieval-augmented LMs only yield marginal gains on reasoning tasks, which can be attributed to weaknesses in both the retrieval and LM components~\cite{behnamghader-etal-2023-retriever,lin2023radit}. 
We hypothesize that this stems from a misalignment between conventional retrieval and LM training objectives, as well as the used datastore.  
Consider answering a factual knowledge-based question: a retriever can efficiently search documents akin to a query in Wikipedia, and an LM can subsequently copy or paraphrase the retrieved information. 
However, the types of beneficial text vary significantly based on the task. {
Existing retrieval systems evaluate the relevance of documents primarily by assessing their high lexical or semantic similarities to the input. 
Yet, such ``relevant'' documents often do not help tasks in reasoning or general language understanding~\cite{rubin-etal-2022-learning}. }
It is still unclear what makes certain retrieved contexts more effective than others. 
The heavy dependence on Wikipedia as a datastore (Section~\ref{sec:survey_applications}) could also limit its effectiveness, as real-world applications frequently encounter queries that may not find direct answers in Wikipedia~\cite{asai-choi-2021-challenges}. 

\vspace{-0.2cm}
\paragraph{\hlc{C2: Limited interactions between retrievers and LMs}.} 
Common approaches, such as RAG, often straightforwardly entail appending retrieved results to the input of pre-trained parametric LMs and adopting input augmentation (Section~\ref{sec:survey_arc}){, due to its simplicity and effectiveness by leveraging state-of-the-art parametric LMs. }
However, these methods lack close interactions between the retrieval and LM components throughout both training and inference. 
This deficiency amplifies issues such as unsupported generations~\cite{gao2023enabling} or susceptibility to irrelevant context, as noted in~\citet{Yoran2023MakingRL,shi2023large}.
Moreover, input augmentation increases the context length of LMs, leading to an exponential increase in inference costs~\cite{xu2023recomp}. 
This becomes particularly problematic when downstream applications require systems to assimilate information from multiple documents~\cite{fan-etal-2019-eli5}. 
Extended context can also induce LMs to overlook significant portions of the input~\cite{liu-etal:2023:tacl}. 

\vspace{-0.2cm}
\paragraph{\hlc{C3: Lack of infrastructure specialized for retrieval-based LMs}.}
Relative to parametric LMs, the optimization of retrieval-augmented LM training procedures has been comparatively under-studied, from both methodological and infrastructural standpoints. 
For instance, open-sourced software such as 
PyTorch FSDP\footnote{\url{https://pytorch.org/docs/stable/fsdp.html}} or DeepSpeed\footnote{\url{https://github.com/microsoft/DeepSpeed}} enable resource-efficient parametric LM pre-training via techniques such as Fully Sharded Data Parallelism~\cite{10.14778/3611540.3611569} or Zero Redundancy Optimizers~\cite{deepspeed}, respectively. 
While retrieval-augmented LMs can certainly leverage improvements made to their parametric components, what remains lacking are focused efforts that address challenges unique to retrieval-augmented LMs.
Synchronously updating large-scale indexes during training introduces significant computational overhead, and how to efficiently update the index under normal computational environments remains challenging (Section~\ref{sec:survey_training}). 

Inference in retrieval-augmented LMs can also be significantly more expensive than in standard parametric LMs~\cite{mallen2022not}, especially if the datastore is large (e.g., over one trillion tokens). 
As scaling pre-training data leads to better parametric LMs, some studies empirically show that scaling the datastoresis promising~\cite{borgeaud2021improving}. 
Yet, nearest neighbor searches over billions of embeddings without extensive tricks can consume hundreds of GPUs or prohibitively high RAM usage. 
Scaling costs thus hinder prior efforts to use larger datastores   (Section~\ref{sec:survey_applications}).

%% file: sections/sec4_challenges.tex
{
We believe that the community needs to develop robust intelligent systems based on retrieval-augmented LMs that surpass fully parametric LMs. 
Here, we present a roadmap to overcome the technical constraints associated with retrieval-augmented LMs discussed in Section~\ref{sec:limitations}. 
}

\subsection{Rethinking Retrieval and the Datastore (\hlc{\bf C1}) }
\label{sec:philosophy}

\paragraph{Beyond semantic and lexical similarity.}
Extending the use of retrieval-augmented LMs beyond conventional knowledge-centric tasks necessitates the formulation of a new definition for ``relevance'' of the input query and documents in the datastore. 
This is essential for excelling in tasks as in those tasks informative text may not exhibit semantic or lexical similarity to the input query.
{
Recent works show that few-shot in-context learning demonstrations~\cite{su2022selective} or even unlabeled text~\cite{lyu-etal-2023-z} could boost model performance on reasoning or language understanding tasks. Yet, what makes certain documents helpful (e.g., underlying reasoning patterns, or writing style) remains an open question. 
Acquiring a better understanding of the characteristics of helpful documents could unlock the potential of retrieval-augmented LMs. 
Furthermore, built upon such understanding, we should build retrieval systems capable of contextualized retrieval, rather than building task-specific retrieval pipelines: {developing a versatile retriever that adjusts its search behavior based on diverse notions of similarity with additional input. }
For instance, instruction-tuned retrievers~\cite{asai-etal-2023-task,su-etal-2023-one} exemplify this direction. 

\vspace{0.2cm}
\noindent{\bf Reconsidering and improving the datastore.}
{
When it comes to wider, general downstream applications, or conversely more expert-domain tasks, over-reliance on a single, general-domain corpus such as Wikipedia may hinder the capability of retrieval-augmented LMs. 
As discussed in Section~\ref{sec:survey_applications}, the curation and composition of the datastore significantly impact the final performance. 
Yet, many open questions exist regarding how to build and ensure high-quality and effective datastores. 
For instance, should we introduce a quality filter to the documents in the datastore, as common practice in pre-training data processing~\cite{black2022gpt}? 
How should we balance multiple domains in a datastore~\cite{shao2023retrievalbased}? 
Despite the abundance of literature on what constitutes good LM pre-training data~\cite{longpre2023pretrainer}, there have been limited explorations so far on what data ought to go into the datastore.  
}

\subsection{Enhancing Retriever-LM Interactions (\hlc{\bf C2})}
\label{sec:challenges_arc}

\paragraph{New architectures beyond input augmentation.}
As discussed, the input augmentation of powerful LMs (e.g., RAG) comes with several limitations that could be addressed by more specialized, integrated architectures, such as output interpolation or intermediate fusion. 
While recent work shows the success of new architectures~\cite{wang-etal-2023-shall,min2022nonparametric,Lan2023CopyIA}, compared to massively pre-trained parametric LMs, their training and model size are often smaller, due to high computational costs for pre-training. 
Furthermore, approaches that employ a smaller granularity of retrieval (e.g., token level in Section~\ref{sec:survey_arc}) pose significant challenges for scaling. 
{We urge collaborative efforts for scalable, effective architecture designs and pre-training---While pre-training retrieval-augmented LMs is computationally expensive, we hope that we can address that challenge through collaborative multi-institution efforts, as in several successful parametric LM pre-training~\cite{workshop2022bloom,groeneveld2024olmo}.}
Recently, \citet{muennighoff2024generative} have introduced generative representational instruction tuning to train a single model for both retrieval and generative tasks, which allows for significantly reducing the latency of RAG by caching representations.

\paragraph{Incorporating retrieval during {LM} pre-training.} 
Off-the-shelf parametric LMs trained without retrieval components often struggle with leveraging additional context~\cite{shi2023large}. 
Pre-training LMs with retrieval has proven to be effective ~\cite{guu2020retrieval,lewis2019bart,izacard2022few}, but often requires significant additional training costs, or non-trivial modifications to the standard LM architecture. 
Recently, \citet{shi2023context} shows that retrieving similar text chunks and reordering pre-training corpora can enhance LMs' abilities to reason over long sequences or perform retrieval augmentation for diverse tasks. 
These improvements do not require the modification of pre-training pipelines or model architectures. 
As such, the exploration of methods to induce LMs to leverage retrieved context with minimal or no additional costs remains promising. 

\paragraph{Further adaptation after pre-training.} 
Significant architecture modification or pre-training are efforts that require massive computing. One promising avenue under resource-constrained environments is to explore adaptations of retrieval-augmented LMs after pre-training. 
For instance, despite the rapid developments of versatile instruction-following parametric LMs, the exploration of instruction-following retrieval-augmented LMs~\cite{lin2023radit,luo2023sail,asai2023self} or RLHF for retrieval-augmented LMs~\cite{nakano2021webgpt,bohnet2022attributed} remains comparatively scarce. 
{Augmenting existing instruction-tuned LMs trained without retrieval can often cause suboptimal performance as the LMs are not explicitly trained to use the retrieved context. }
Further investigation for better post-hoc adaptation recipes (e.g., instruction-tuning, RLHF) for retrieval-augmented LMs may unleash their effectiveness across diverse downstream adaptations. 
Recent studies demonstrate the promise of incorporating additional components to filter out irrelevant context~\cite{xu2023recomp, Yoran2023MakingRL} or instructing an LM to learn to distinguish~\cite{asai2023self}. 
Exploring enhanced pipelines or inference-time algorithms could further improve reliability.

\paragraph{Efficient end-to-end training of retrieval-augmented LMs.}
Retrieval errors often stand out as prominent issues in retrieval-augmented LMs~\cite{asai-choi-2021-challenges, Yoran2023MakingRL}. 
Rather than focusing on optimizing the LM component in isolation, it is crucial to jointly optimize the retriever component. Some tasks have demonstrated success in updating only the input encoding component without modifying the index after pre-training~\cite{izacard2022few, lin2023radit}. Another alternative strategy involves introducing additional components, such as reranking models, and training them in an end-to-end fashion with LMs. 
In many downstream tasks, no supervised labels are available to train retrieval systems. 
Studying effective training strategies without supervision on the latent variable for positively retrieved context~\cite{lee-etal-2019-latent, NEURIPS2021_da3fde15} is essential for enabling the training of retrieval-augmented LMs for a broader range of applications.

\vspace{-0.2cm}
\subsection{Building Better Systems and Infrastructures for Scaling and Adaptation (\hlc{\bf C3})}
\label{sec:challenges_datastore}

\paragraph{Scalable search for massive-scale datastores. } 
We believe significant efforts and expertise from interdisciplinary areas, including systems and algorithms, will enable practitioners to leverage large-scale datasets.   
For instance, exploring compression and quantization algorithms for billions of text embeddings is an important area~\cite{douze2024faiss}, as well as faster nearest neighbor search algorithms~\cite{10.14778/3476249.3476255}. 
Open-sourced toolkits such as FAISS~\cite{johnson2017billion} could accelerate such progress. 
Another bottleneck of datastore-scaling is the storage requirements for millions or billions of encoded documents, and how to efficiently load them during inference. 
Some recent works propose to significantly reduce the index size by storing the index as binary vectors~\cite{yamada-etal-2021-efficient,cao2023btr}. 
Besides algorithmic improvements and system development, another promising avenue is the development of specialized hardware for retrieval-augmented LMs. 
Compared to parametric LMs, retrieval-augmented LMs may require fewer GPUs, while it is often CPU-heavy and requires fast access to the datastore. 
Collaborative efforts from hardware, systems, and algorithms to LM applications could help us tackle these challenging problems. 

\noindent{\bf Standardization and open-source developments.}
There are several repositories such as LangChain,\footnote{\url{https://python.langchain.com/docs/get_started/introduction}} LlamaIndex,\footnote{\url{https://www.llamaindex.ai/}} and DSPy~\cite{khattab2024dspy}\footnote{\url{https://github.com/stanfordnlp/dspy}} that enable practitioners to build RAG on top of existing retrievers, parametric LMs, and user-provided datastores. 
Yet, we still lack a standardized implementation of retrieval-augmented LM pipelines and evaluation benchmarks that can flexibly accommodate a range of architectures and training configurations (Sections \ref{sec:survey_arc} and \ref{sec:survey_training}) beyond RAG.  
As open-sourced efforts have facilitated the rapid progress of parametric LMs, we urge the community to similarly build a standardized open-source implementation for retrieval-augmented LMs.

%% file: sections/sec5_conclusion.tex
This paper advocates for retrieval-augmented LMs as the next generation of LMs to build more reliable, adaptable, and attributable intelligent systems. 
{Despite their notable advantages over parametric LMs, their adoption remains limited. This limitation may be attributed to the focus on a narrow form of retrieval augmentation, which simply combines exiting retrieval models and LMs in post-hoc manners to supplement parametric LMs.} 
{We outline a roadmap for fundamentally advancing retrieval-augmented LMs in terms of architectures, training methodologies, and infrastructure. 
We emphasize the importance of collaborative interdisciplinary efforts to achieve these advancements.}

\section*{Impact Statements}

{
We believe the adoption of retrieval-augmented LMs could address those fundamental limitations inherent to parametric LMs. 
We hope that this position paper will inspire further exploration in these areas, and collaboratively foster the advancement of retrieval-augmented LMs.
However, concerns may arise. The effectiveness of retrieval-augmented LMs in tasks beyond knowledge-intensive domains remains an open question, necessitating thorough assessments. Furthermore, retrieval-augmented LMs may not completely address issues such as hallucinations. 
}

%% file: sections/appendix.tex
\section{Progress of Parametric LMs}
\paragraph{The rise of parametric LMs.}
 
Pre-training to develop better parametric representations of text has been recently extensively studied. 
BERT~\cite{devlin-etal-2019-bert} is considered to be the first pre-trained LM trained on large-scale text, built upon prior great success on pre-trained contextualized representations (ELMo; ~\citealt{peters-etal-2018-deep}). 
BERT is an encoder-only, masked LM that is trained to fill in blanks (masked tokens) during pre-training, similar to several widely used pre-trained models such as RoBERTa~\cite{liu2020roberta}. 
BART~\cite{lewis2019bart} or T5~\cite{roberts2019exploring}, on the other hand, are encoder-decoder models that are trained in both masked and autoregressive manners. 
GPT~\cite{radford2018improving} and GPT-2~\cite{Radford2019LanguageMA} are decoder-only, autoregressive LMs that predict continuations of input tokens. 
Recent research highlights the advantages of expanding both the parameter count of models and the scale of pre-training datasets~\cite{Rae2021ScalingLM}. 
Many proprietary LLMs such as 175B GPT-3~\cite{black2022gpt}, GPT-4~\cite{openai2023gpt} or publicly released checkpoints such as Llama 1~\cite{touvron2023llama1} and Llama 2~\cite{touvron2023llama2}, which training a smaller number of parameters on trillions of tokens, have shown strong performance across various tasks.

\paragraph{{Versatile, instruction-following systems}.}
Starting with GPT-3~\cite{NEURIPS2020_1457c0d6}, large parametric LMs have demonstrated an emergent ability known as {\it in-context learning}---the ability to adapt to new tasks through few-shot prompting without needing any updates to its parameters. 
Further studies demonstrate the impact of large-scale supervised training across varied input-output pairs, as well as subsequent refinements using reinforcement learning with human feedback (RLHF), resulting in powerful instruction-following models~\cite{ouyang2022training,wang2023far,dubois2023alpacafarm}.

\paragraph{{Infrastructure for scalability and efficiency}.} 
The necessity of training and hosting massive parametric LMs has motivated active interdisciplinary research and open-source developments to reduce the computational costs and time of training and inference. 
For instance, open-sourced software such as 
PyTorch FSDP\footnote{\url{https://pytorch.org/docs/stable/fsdp.html}} or DeepSpeed\footnote{\url{https://github.com/microsoft/DeepSpeed}} enable more resource-efficient parametric LM pre-training via techniques such as Fully Sharded Data Parallelism~\cite{10.14778/3611540.3611569} or Zero Redundancy Optimizers~\cite{deepspeed}, respectively.   
FlashAttention~\cite{dao2022flashattention} accelerates training and long-context processing.  
Intensive ongoing research addresses the challenges of high inference costs of massive parametric LMs; memory-efficient inference algorithms such as PagedAttnetion~\cite{kwon2023efficient} used in vllm\footnote{\url{https://github.com/vllm-project/vllm}} are proposed to speed up the inference of billion-scale parametric LMs.

\section{Detailed Taxonomy of Retrieval-augmented LMs}
\label{app_sec_survey}
\subsection{Architectures}
\label{app_sec_survey_arc}

We introduce a taxonomy of architectures of retrieval-based LMs. 
Our taxonomy (Figure~\ref{fig:fig2}) is based on three axes: (1) the {\bf granularity of retrieval} (what to retrieve), (2) the {\bf incorporation method} (how to use retrieval), and the (3) {\bf frequency of retrieval} (when to retrieve).  
This taxonomy extends the summarized taxonomy in Section~\ref{sec:survey_arc}.

\subsubsection{Granularity of Retrieval}
\label{sec:granu}
We specify the retrieval granularity as follows: text chunks or smaller granularity such as tokens, phrases, or entities. 
While it has shown to be effective, text chunks often contain more information than necessary, resulting in redundancy.  

\noindent {\bf Text chunks.}
The retrieval of text chunks, such as 100-word paragraphs, is a prevalent strategy in widely used retrieval-augmented LMs such as REALM, RAG, and RETRO. 
To implement this, a large-scale corpus $\mathcal{D}$ is segmented into text chunks based on the number of tokens or predefined structures like section headers or paragraphs. 
Retrieved chunks are typically integrated into input space or intermediate layers, which we discuss in detail in the following section, while recent work shows that the choice of length significantly affects performance~\cite{chen2023dense}. 
LMs are expected to predict output token probability distributions by jointly leveraging their original knowledge in parameters and retrieved text chunks.

\noindent {\bf Tokens and phrases.}
Several work explores much smaller units such as  tokens~\cite{khandelwal2019generalization} or phrases~\cite{min2022nonparametric}. 
Given the input prompt $x$, such token or phrase retrieval-augmented LMs directly search possible next tokens from the datastore by matching the input prompt and similar prefixes in the datastore, instead of making the LM read and generate from the vocabulary.
Token or phrase retrieval can often result in a much larger index size compared to text chunk retrieval given the same size of datastore (i.e., the number of embeddings is by default equal to the number of tokens in the datastore). 

\subsubsection{Incorporation Method}
\label{sec:incorporation}
Another important axis is how the retrieved information. 
Essentially, retrieval-augmented LMs' architectures can be classified into the following three groups: 1)  {\bf input augmentation}, 2) {\bf intermediate fusion}, and ) {\bf output interpolation}. 

\noindent {\bf Input augmentation.}
Input augmentation simply augments the original input $x$ with retrieved results $z$ in the input space of the LM $\theta$ and runs a standard LM inference. 
As in the pioneering work by \citet{chen-etal-2017-reading}, input augmentation enables flexible plug-ins of different models for retrieval and LM components. 
For instance, ATLAS~\cite{izacard2022few} and REALM pre-trains LMs jointly with the retriever, while some recent work leverage off-the-shelf pre-trained LMs and retrievers~\cite{ram2023context,shi2023replug}. 
One notable bottleneck is its redundancy and inefficiency; encoding many documents together in input space faces context length window limitations and increases inference costs exponentially~\cite{xu2023recomp}. 
While some work such as FiD~\cite{izacard2022few} 
explores parallel encoding to overcome such inefficiencies, still the same documents need to be encoded repeatedly for each input $x$. 

\noindent {\bf Intermediate fusion.}
To integrate retrieved results in a more scalable manner, RETRO~\cite{borgeaud2021improving} introduces a new attention mechanism called chunked cross attention (CCA). CCA takes many pre-encoded text chunks, which are independent of query $x$ unlike input augmentation, simultaneously in intermediate spaces by adding a new block between standard cross attention and feed-forward network in Transformer~\cite{NIPS2017_3f5ee243}. 
Recently, RETRO++~\cite{wang-etal-2023-shall} and InstructRetro~\cite{anonymous2023instructretro} incorporated CCA into powerful autoregressive LMs. 
However, a drawback of intermediate fusion is the need for architecture modification and pre-training of LMs for the new encoding blocks, potentially limiting widespread adoption. 
Several studies focus on similar architectures for retrieving information from long-context input~\cite{wu2022memorizing,rubin2023long}.

\noindent {\bf Output interpolation.}
The two incorporation methods described above still let LMs generate continuations from their vocabularies, which often results in unsupported or unattributed generations~\cite{liu-etal:2023:tacl,gao2023enabling,bohnet2022attributed}.
Instead, some models directly manipulate output token distributions. 
kNN LM interpolates the original LM's softmax token distributions with retrieved token distribution without additional training. 
Some work extends this direction by designing new training objectives~\cite{zhong2022training} or completely replacing a nonparametric
distribution over every phrase in a reference
corpus~\cite{min2022nonparametric,Lan2023CopyIA}.

\subsubsection{Frequency of Retrieval}
\label{sec:frequency}
Another significant design choice in retrieval-augmented LMs is the frequency of retrieval. 
In essence, opting for more frequent retrieval tends to enhance performance, but comes at the expense of increased computational overhead. 
Retrieving once before generating given input $x$ has been widely used such as REALM or DrQA, often in input space incorporation architectures. 
kNN LM, on the other hand, retrieves at every token, or some work retrieves every $k$ token to maintain the relevance between the target sequence and retrieved context~\cite{ram2023context}. 
Several recent papers introduce methods that make LMs adaptively decide when to retrieve~\cite{jiang2023active,asai2023self}.

\subsection{Applications and Datastore} 
\label{app_sec:applications_datastore}
This section briefly reviews the applications of retrieval-augmented LMs and used datastores.

\paragraph{Applications.}
Retrieval-augmented LMs are shown to be effective across a range of NLP tasks, including discriminative and generative tasks. 
The majority of prior work is often evaluated on knowledge-intensive tasks, such as open-domain QA~\cite{kwiatkowski-etal-2019-natural}, fact verification~\cite{thorne-etal-2018-fever} and knowledge-grounding dialogue~\cite{shuster2021retrieval}. 
For such tasks, Wikipedia is often used as the sole knowledge source, while some recent work directly combines LMs with commercial search engine APIs. 
For non-knowledge-intensive tasks, the usage of training instances (labeled data) as the datastore has been widely explored, demonstrating effectiveness on tasks like machine translation~\cite{khandelwal2021nearest,zhong2022training}. 
Some recent works such as kNN-Prompt~\cite{shi2022nearest} or NPM~\cite{min2022nonparametric} leverage larger pre-training corpora (e.g., the Pile; \citealt{gao2020pile}) for more general language understanding tasks (e.g., sentiment analysis) or entity translations. 
\citet{yu-etal-2022-retrieval} build a new large-scale corpus consisting of 20 million commonsense documents collection
from both open-domain knowledge sources. 
Several works on code generations use similar codes~\cite{hayati-etal-2018-retrieval} or documentation~\cite{zhou2023docprompting} of APIs. 
Designing and building a reliable datastore is a key challenge in retrieval-augmented LMs. 
Across those papers, retrieval-augmented LMs have shown significant improvements over parametric LMs. 

{Furthermore, retrieval-augmented LMs have been applied beyond general-domain, English text data. Several works explore retrieving from multilingual data~\cite{NEURIPS2021_3df07fda,nie-etal-2023-cross} or multiple modalities~\cite{yasunaga2022retrieval,chen-etal-2022-murag}---which includes underexplored modalities such as robot controls~\cite{zha2023distilling}. While prior work often explores retrieving from general-domain datastore such as Wikipedia, some recent work shows that retrieving from a targeted datastore is largely helpful to solve more challenging expert domain tasks, such as theorem proving~\cite{welleck2022naturalprover,yang2023leandojo} or molecule generation~\cite{wang2023retrievalbased}.  }